\documentclass[11pt,a4paper]{article}
\usepackage[hyperref]{emnlp2020}
\usepackage{times}
\usepackage{latexsym}

\usepackage{graphicx}
\usepackage{array,multirow}
\usepackage{tikz}
\usepackage{tikz-dependency}

\usepackage{microtype}

\aclfinalcopy 

\newcolumntype{P}{>{\raggedright\arraybackslash}p{\columnwidth}}
\newcolumntype{M}{>{\raggedright\arraybackslash}m{\columnwidth}}

\title{Learning Music Helps You Read: Using Transfer to Study Linguistic Structure in Language Models}

\author{Isabel Papadimitriou \\
Stanford University \\
  \texttt{isabelvp@stanford.edu} \\\And
  Dan Jurafsky \\
  Stanford University \\
  \texttt{jurafsky@stanford.edu} \\
  }

\date{}

\begin{document}
\maketitle
\begin{abstract}
We propose transfer learning as a method for analyzing the encoding of grammatical structure in neural language models. We train LSTMs on non-linguistic data and evaluate their performance on natural language to assess \textit{which kinds of data induce generalizable structural features} that LSTMs can use for natural language.  We find that training on non-linguistic data with latent structure (MIDI music or Java code) improves test performance on natural language, despite no overlap in surface form or vocabulary. To pinpoint the kinds of abstract structure that models may be encoding to lead to this improvement, we run similar experiments with two artificial parentheses languages: one which has a hierarchical recursive structure, and a control which has paired tokens but no recursion. Surprisingly, training a model on either of these artificial languages leads to the same substantial gains when testing on natural language. Further experiments on transfer between natural languages controlling for vocabulary overlap show that zero-shot performance on a test language is highly correlated with typological syntactic similarity to the training language, suggesting that representations induced by pre-training correspond to the cross-linguistic syntactic properties. Our results provide insights into the ways that neural models represent abstract syntactic structure, and also about the kind of structural inductive biases which allow for natural language acquisition. \footnote{We release code to construct the corpora and run our experiments at \url{https://github.com/toizzy/tilt-transfer}}
\end{abstract}

\section{Introduction}


Understanding how neural language models learn and represent syntactic structure
is an important analytic question for  NLP. Recent work  has directly probed the internal activations of models \cite{conneaufing, grain, johnprobe, urvashibert}, or fed them curated inputs that depend on complex syntax \cite{linzen16, gulordava, olmpics, mccoy2020}, in order to uncover latent syntactic awareness.

\begin{figure}[t]
\includegraphics[width=0.95\columnwidth]{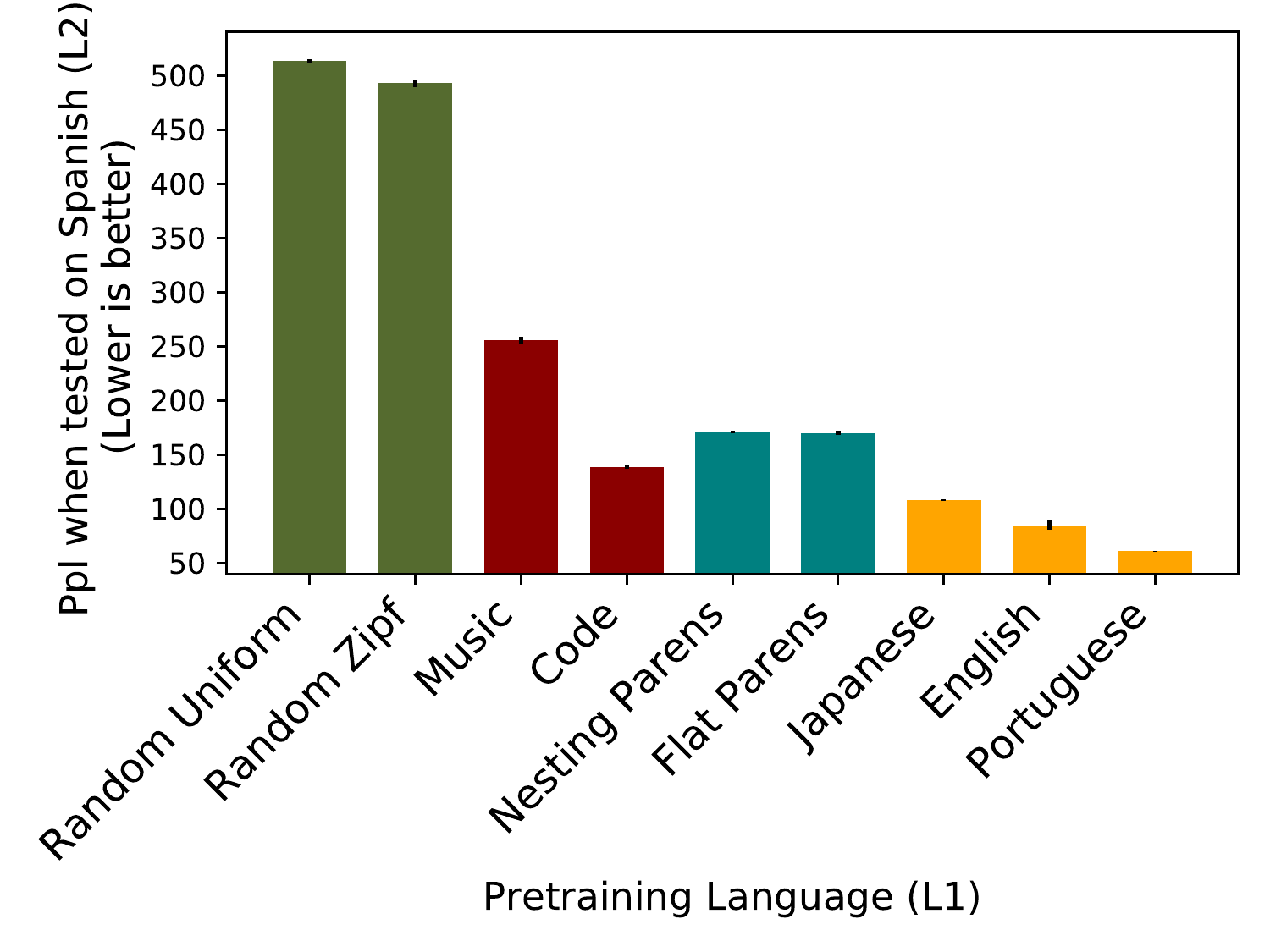} \caption{We find that LSTM LMs can utilize various types of non-linguistic structure to help learn to model human language, and that nested hierarchical structure does not lead to more expressive encodings than flat, head-dependency pair structure. We also find that LSTM LMs learn representations that correlate with typological syntactic feature distance, allowing them to  transfer more effectively from languages which are grammatically similar.}
\end{figure}


\begin{figure*}[t]
\centering
\includegraphics[width=0.8\textwidth,trim={0 2cm 0 0.5cm},clip]{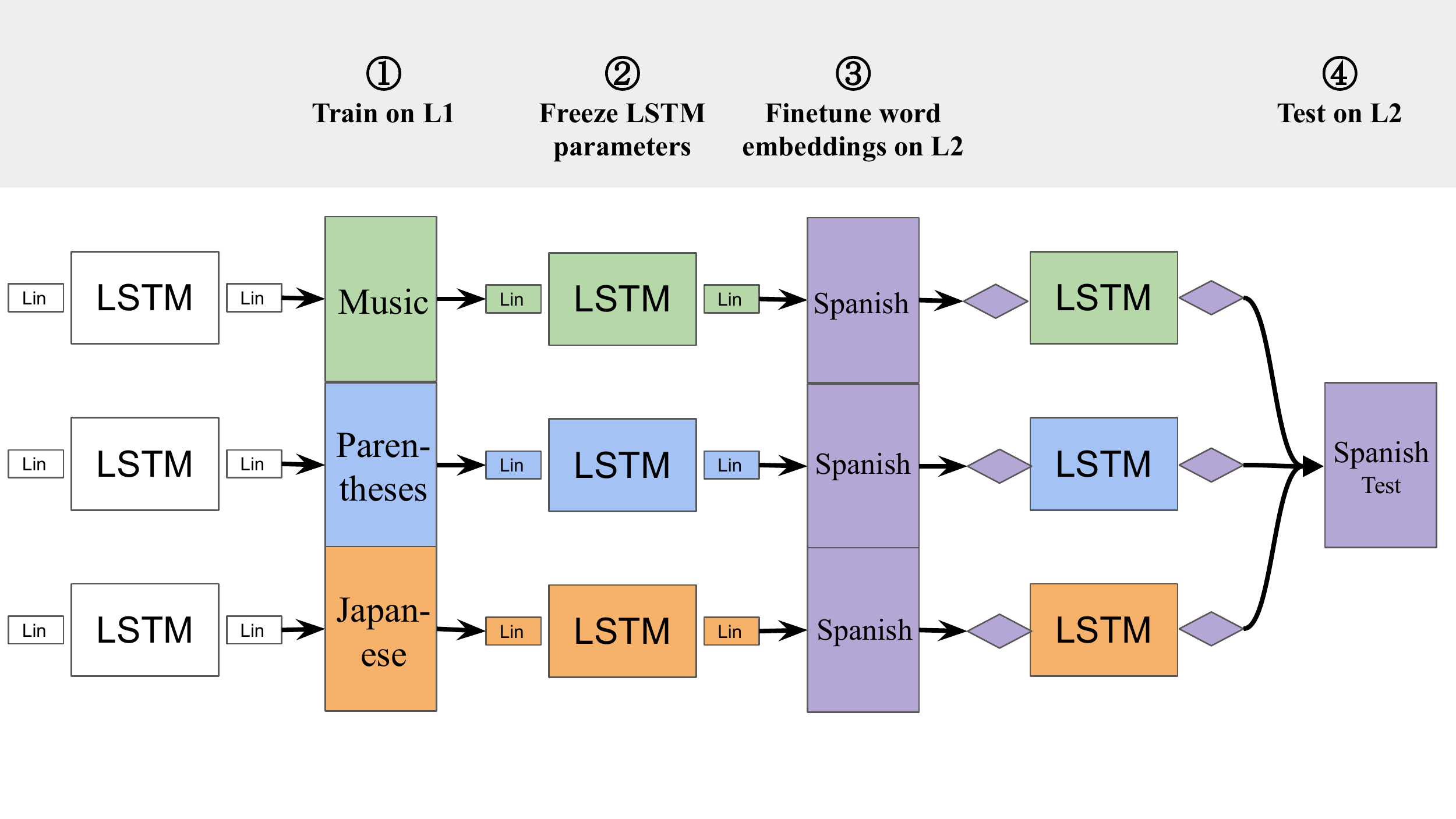}
\caption{Diagram illustrating our training procedure: $k$ models are trained on $k$ L1 languages, and then their LSTM weights are frozen while their linear layers are finetuned on a common L2 language (in our case, we always use Spanish as the L2). We can then compare their performance on the common L2.}
\label{trainingdiagram}
\end{figure*}

We propose a different approach: we measure the structural awareness  of a language model 
\textit{by studying how much this structure acts  as an inductive bias to improve learning when we transfer from one language or symbolic system to another}.

We train LSTM models on data with varying degrees of language-like structure (music, Java code, nested symbols), and then evaluate their performance on natural language. Before evaluation, we freeze the LSTM parameters and fine-tune the word embeddings on the evaluation language. This lets us see if the training data induces language-like structure in the recurrent parameters of LSTMs--- despite removing vocabulary-level confounders. By assessing if representations are useful \textit{across} languages, we examine the generalizable representations of grammar that LSTMs encode. We call this new method the Test for Inductive Bias via Language Model Transfer (TILT).

Firstly, we examine the transfer of abstract structural features from languages that are very different on the surface from human language. We find that pretraining an LSTM on music data\footnote{We use the MAESTRO music dataset, which utilizes an exact symbolic representation of music (like a music score) that is sequentialized for sequence modelling} or Java code greatly improves transfer to human language over pretraining on structureless random baseline data. To test if the gain in performance is due to the LSTM utilizing the recursive nature of music and code, we train models on an artificial language with recursion (hierarchically nested symbols) and observe that they also perform well when evaluated on human language. However, we also surprisingly find that recursion is a sufficient, but not necessary condition for generalizable, language-like grammar induction. We observe similar gains when pretraining on a language of matching pairs that do not nest hierarchically, showcasing the importance of non-hierarchical head-dependent-type relations in LSTM language processing.

Lastly, in transfer experiments between different human languages, we find that transfer is better between languages that are syntactically typologically similar, even with no vocabulary overlap. This suggests that models have the ability to form representations of typologically sensible properties rather than relying on ad-hoc or non-natural representations.  For this result we draw on recent interlingual work such as \citet{artetxezero}, \citet{edo-zero}, and \citet{xnli}, extending it to use typological distance to turn these observations into quantitative probes.

The TILT method allows us to ask a complementary set of questions to those answered by current analysis methods. TILTs demonstrate the abstract structural notions that LSTMs can learn, rather than probing for the manifestation of a particular known structure, as in most current methods. By examining the pretraining structures that give LSTMs a better ability to model language, we also contribute to the more general cognitive question of what structural inductive biases a learner needs to be able to easily acquire human language.

\section{Architecture and Training}

Our methodology consists of training LSTM language models on $k$ different first languages (L1s) which include natural languages, artificial languages, and non-linguistic symbol systems, and testing the performance of these models on a common second (L2) language. In our case, we used Spanish as the common L2. Before testing on the L2 test set, we fine-tune the linear embedding layer of the models on the L2 training set, while keeping the LSTM weights frozen. This aligns the vocabulary of each model to the new language, but does not let it learn any structural information about the L2 language. Though word embeddings do contain some grammatical information like part of speech, they do not contain information about how to connect tokens to each other -- that information is only captured in the LSTM. Figure \ref{trainingdiagram} illustrates our training process. \footnote{All pretraining jobs took less than 2 days to run on one GPU, all finetuning jobs took less than 1 day to run on one GPU.}

We vary the L1 languages and maintain a common L2 (instead of the other way around) in order to have a common basis for comparison: all of the models are tested on the same L2 test set, and therefore we can compare the perplexity scores. We run $n=5$ trials of every experiment with different random seeds.  
Any high-resource human language would have provided a good common L2, and Spanish  works well for our human languages experiments due to the fact that many higher-resource languages fall on a smooth gradation of typological distance from it (see Table \ref{walstable}). 

We use the AWD-LM model \cite{awdlm} with the default parameters of 3 LSTM layers, 300-dimensional word embeddings, a hidden size of 1,150 per layer, dropout of 0.65 for the word embedding matrices and dropout of 0.3 for the LSTM parameters. We used SGD and trained to convergence, starting the learning rate at the default of 30 and reducing it at loss plateau 5 times.

Much of the work on multilingual transfer learning has speculated that successes in the field may be due to vocabulary overlap (see for example \citet{beto}). Since our work focuses mostly on syntax, we wanted to remove this possibility. As such, we shuffle each word-to-index mapping to use disjoint vocabularies for all languages: the English word ``Chile'' and the Spanish word ``Chile'' would map to different integers. This addresses the confound of vocabulary overlap, as all language pairs have zero words in common from the point of view of the model. 

Since the vocabularies are totally separated between languages, we align the vocabularies for all L1-L2 pairs by finetuning the word embeddings of all the pretrained models on the Spanish (L2) training data, keeping the LSTM weights frozen. By doing this, we remove the confound that would arise should one language's vocabulary randomly happen to be more aligned with Spanish than another's.
These controls ensure that lexical features, whether they be shared vocabulary or alignment of randomly aligned indices, do not interfere with the experimental results which are meant to compare higher-level syntactic awareness. 

\section{Experiment 1: Random Baselines} \label{randomsec}

We run our method on a random baseline L1: a corpus where words are sampled uniformly at random. This gives us a baseline for how much information we gain finetuning the word embeddings to the L2, when there has not been any structurally biasing input to the LSTM from the L1. 

We also examine the importance of vocabulary distribution by training on a random corpus that is sampled from a Zipfian distribution. Human languages are surprisingly consistent in sharing a roughly Zipfian vocabulary distribution, and we test how pretraining on this distribution affects the ability to model human language. \footnote{See \citet{piantadosizipf} for a  review of cognitive, communication and memory-based theories seeking to explain the ubiquity of power law distributions in language.}

    \begin{figure*}[t]
    \small
    \begin{tabular}{|P|p{\columnwidth}|}
    \hline
    \underline{\textbf{Random}}&
    \vspace{0.008cm}
    \multirow{4}{\columnwidth}{
    The random corpora are sampled randomly from the Spanish vocabulary. There is no underlying structure of any kind that links words with each other. All words are equally likely to be sampled in the Uniform corpus, while common words are more likely in the Zipfian corpus.}\\
    \textbf{Uniform:} \texttt{marroqu\'{i}n jemer pertenecer osasuna formaron citoesqueleto relativismo}&\\
    \cline{1-1} \vspace{0.005cm}\textbf{Zipf:} \texttt{en con conocidas y en los victoriano como trabajar} $\langle$unk$\rangle$ \texttt{monte * en juegos d\'{i}as en el}\vspace{0.1cm}& \\
    \hline
    \underline{\textbf{Music}}\hspace{0.5cm}\raisebox{-\totalheight}{\includegraphics[height=2cm, width=0.7\columnwidth]{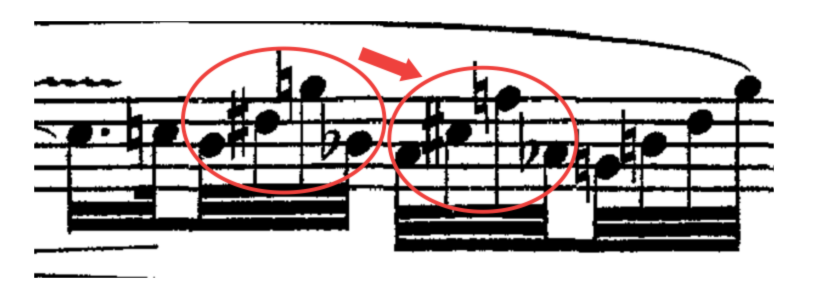}}&
    \vspace{0.008cm}
    The music data is encoded from classical piano performances according to the MAESTRO standard. Music is structured on many levels. The red arrow in the example illustrates how, on a small timescale, each note is linked to its corresponding note when a motif is repeated but modulated down a whole-step.\\
    \hline
    \underline{\textbf{Code}}
    \begin{verbatim}
    if (coordFactor == 1.0f) 
        return sumExpl
    else {
        result = sum * coordFactor 
    } \end{verbatim}&
    \vspace{0.008cm}
    The code corpus is composed of Java code. The above snippet demonstrates some kinds of structure that are present in code: brackets are linked to their pairs, \texttt{else} statements are linked to an \texttt{if} statement, and coreference of variable names is unambiguous.
    \\
    \hline
    \underline{\textbf{Parentheses}}&
    \vspace{0.008cm}
    \multirow{3}{\columnwidth}{
    Our artificial corpora consist of pairs of matching integers. In the Nesting Parentheses corpus, integer pairs nest hierarchically and so the arcs do not cross. In the Flat Parentheses corpus, each integer pair is placed independently of all the others, and so the arcs can cross multiple times.
    \\
    ~\\(There is a one-to-one mapping between Spanish words and integers and so 
    these integers are sampled from the same Spanish vocabulary distribution as the Random Zipfian corpus. 
    We visualize these corpora here with integers and the Random corpora with words for simplicity).
    }\\
    \textbf{Nesting:}
    \raisebox{-\totalheight}{\begin{dependency}[edge unit distance=0.8ex,label theme = simple, edge theme = iron]
      \begin{deptext}[column sep=0.09cm]
        0 \& 29 \& 29 \& 0 \& 0 \& 5 \& 5 \& 0 \& 1016 \& 1016 \& 9 \& 8 \& 8\& 28 \& 28 \& 9\\
      \end{deptext}
      \depedge{1}{4}{}
      \depedge{2}{3}{}
      \depedge{5}{8}{}
      \depedge{6}{7}{}
      \depedge{9}{10}{}
      \depedge{11}{16}{}
      \depedge{12}{13}{}
      \depedge{14}{15}{}
    \end{dependency}}& \\
    \cline{1-1}\vspace{0.005cm}\textbf{Flat:}\raisebox{-\totalheight}{\begin{dependency}[label theme = simple, edge theme = iron, edge unit distance=1ex]
      \begin{deptext}[column sep=0.09em]
      21 \& 13 \& 21 \& 6294 \& 13 \& 6294 \& 5 \& 5471 \& 5 \& 32 \& 32 \& 5471\\
      \end{deptext}
      \depedge{1}{3}{}
      \depedge{2}{5}{}
      \depedge{4}{6}{}
      \depedge{7}{9}{}
      \depedge{8}{12}{}
      \depedge{10}{11}{}
    \end{dependency}}& \\
    \hline
    \end{tabular}
    \caption{Examples illustrating the content of our non-linguistic corpora for Experiments 1-3. All examples are taken from the corpora.}
    \label{nonlingexamples}
    \end{figure*}

\subsection{Data}

Our random corpora are sampled from the Spanish vocabulary, since Spanish is the common L2 language across all experiments. Words are sampled uniformly for the Uniform Random corpus, and drawn from the empirical Spanish unigram distribution (as calculated from our Spanish training corpus) for the  Zipfian Random corpus. Illustrative examples from all of our corpora can be found in Figure \ref{nonlingexamples}. The random corpora are controlled to 100 million tokens in length.

    \begin{figure}[t]
    \includegraphics[width=\columnwidth]{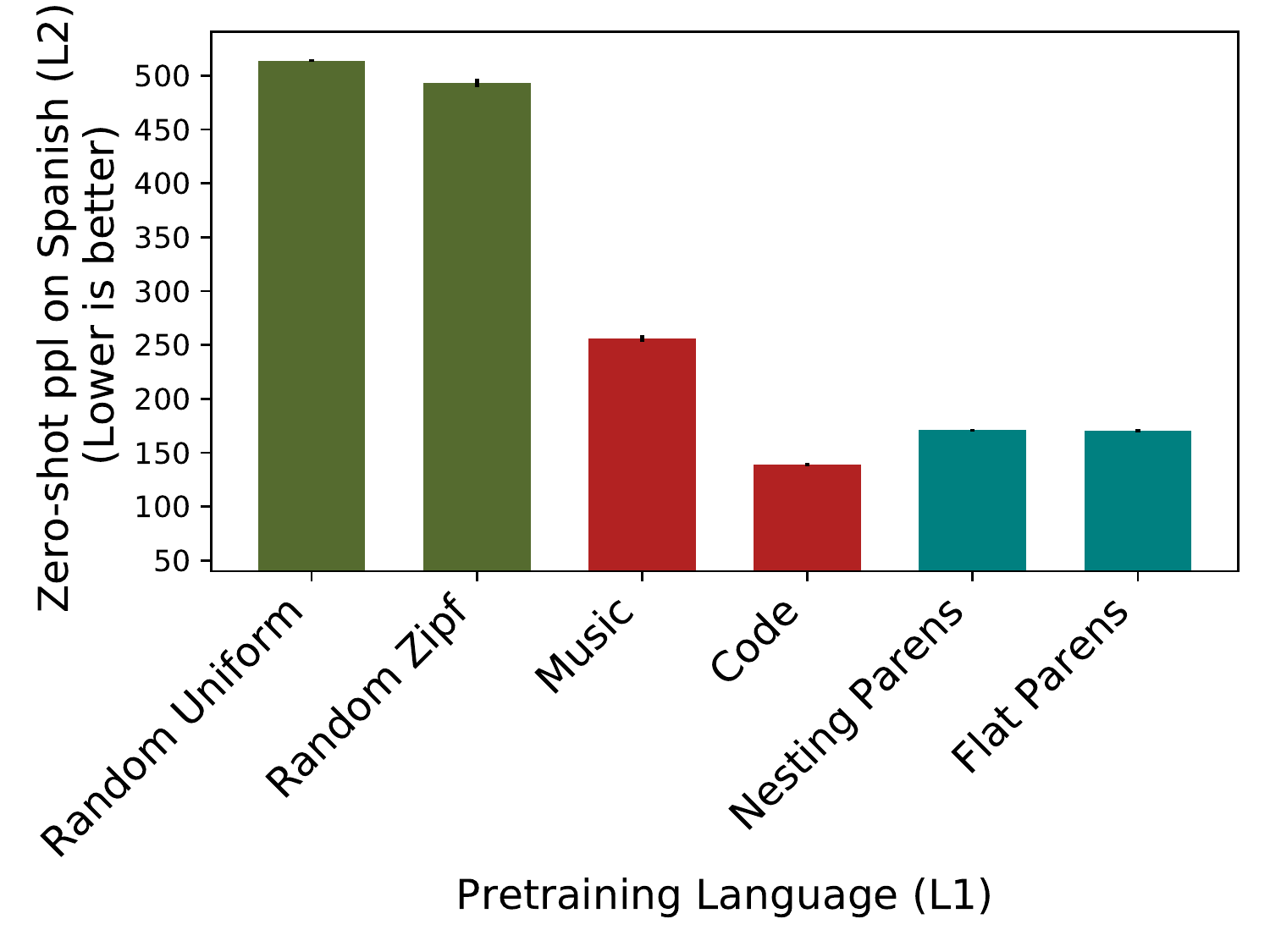}
    \caption{Results of Experiments 1 through 3, training on non-linguistic corpora. Error bars on all bars indicate a 95\% $t$-test confidence interval over 5 restarts with different random seeds. All structured data is much better to train on than random data, including music which has a totally divergent vocabulary surface form from the rest. The two parentheses corpora result in equivalent perplexities, even though one has a hierarchical underlying structure and the other does not.}
    \label{nonlingresults}
    \end{figure}

\subsection{Results}

When tested on Spanish, the average perplexity is 513.66 for models trained on the Random Uniform corpus and 493.15 for those trained on the Random Zipfian corpus, as shown in Figure \ref{nonlingresults}.
These perplexity values are both smaller than the vocabulary size, which indicates that the word embedding finetuning captures information about the test language even when the LSTM has not been trained on any useful data.

The models trained on the Zipfian Random corpus are significantly better than those trained on the Uniform corpus ($p << 0.05$, Welch's $t$-test over $n=5$ trials). However, even though training on a Zipfian corpus provides gains when compared to training on uniformly random data, in absolute terms performance is very low. This indicates that, without higher-level language-like features, there is very little that an LSTM can extract from properties of the vocabulary distribution alone.

The Zipfian Random baseline is controlled for vocabulary distribution: if an experiment yields better results than the Zipfian Random baseline, we cannot attribute its success only to lexical-level similarity to the L2. Therefore, models that are more successful than the Zipfian baseline at transfer to human language would have useful, generalizable syntactic information about the structures that link tokens.

\section{Experiment 2: Non-linguistic structure}

In this experiment, we test the performance of LSTMs on Spanish when they have been trained on music and on code data. While music data especially is very different from human language on the surface level, we know that music and code both contain syntactic elements that are similar to human language.\footnote{See for example \citet{generativemusic} for grammatical structure in music.} By comparing performance to our random baselines, we ask: can LSTMs encode the abstract structural features that these corpora share with natural language in a generalizable way that's usable to model human language? 

\subsection{Data}

For our music data we use the MAESTRO dataset of \citet{maestro}.  The MAESTRO dataset embeds MIDI files of many parallel notes into a linear format suitable for sequence modelling, without losing musical information. The final corpus has a vocabulary of 310 tokens, and encodes over 172 hours of classical piano performances. \footnote{The MAESTRO dataset is available at \url{https://magenta.tensorflow.org/datasets/maestro}}

For programming code data, we used the Habeas corpus released by \citet{habeas}, of tokenized and labelled Java code. \footnote{The Habeas corpus is available at \url{https://github.com/habeascorpus/habeascorpus-data-withComments}} We took out every token that was labelled as a comment so as to not contaminate the code corpus with natural language. 

The music corpus is 23 million tokens in length and the code corpus is 9.5 million. We cannot effectively control the lengths of these corpora to be the same as all of the others, since there is no controlled notion of what one token means in terms of information. However, we only compare these results to the random baseline, which we have trained on 100 million tokens -- if the LSTMs trained on these corpora are under-specified compared to the baseline, this would only strengthen our results.

\subsection{Results}

Our results show that language models pretrained on music are far better at modelling Spanish than those pretrained on random data. As shown in figure \ref{nonlingresults}, LSTMs trained on music data have an average performance of 256.15 ppl on Spanish, compared with 493.15 when training on the Zipfian random corpus. This discrepancy suggests that the model, when training on music, creates representations of the relationships between tokens which are generalizable and can apply to Spanish.

The music corpus is markedly different from the Spanish corpus by most measures. Most saliently, MAESTRO uses a vocabulary of just 310 tokens to encode various aspects of music like volume and note co-occurrence.\footnote{For consistency, the model still has a word embedding matrix of 50,000 rows, but during training only ever sees words 1-310, meaning that much of the word embedding space has never been seen by the LSTM part of the model.} This is in contrast to the Zipfian Random corpus, which has the same surface-level vocabulary and distribution as Spanish, yet models trained on it perform on average 237 ppl worse compared to those trained on the music corpus. Since the surface forms between music and language are so different, the difference in performance cannot be based on surface-level heuristics, and {\em our results suggest the presence of generalizable, structurally-informed representations in LSTM language models}.

We also show that models trained on Java code can transfer this knowledge to a human L2 better than the random baseline. Syntactic properties of code such as recursion are similar to natural language, though code is constructed to be unambiguously parsed and lacks a lot of the subtlety and ambiguity that characterizes natural language. Models trained on code have an average perplexity of 139.10 on the Spanish test set. The large discrepancy between this performance and the baseline indicates that LSTMs trained on code capture the syntactic commonalities between code and natural language in a manner that is usable for modelling natural language.

Our results on non-linguistic data suggest that LSTMs trained on structured data extract representations which can be used to model human languages. The non-linguistic nature of these data suggests that it is something structural about the music and Java code that is helping in the zero-shot task. However, there is a multitude of structural interpretations of music, and it is not clear what kinds of structure the LSTM encodes from music. In the next experiment, we create simple artificial corpora with known underlying structures in order to test how the LMs can represent and utilize these structures.

\section{Experiment 3: Recursive Structure}

In this experiment, we isolate and assess possible structural features of music and code that may explain the results of Experiment 2. 
The most widely-known structural hypothesis is the claim of \citet{hauser2002faculty} that the narrow language faculty in humans (the inductive bias in the mind/brain that allows humans to acquire and develop language) can be reduced to just recursion. Given the prominence of such theories, it is natural to ask: is it the underlying recursive nature of music and code data that causes the gains that we observe in Experiment 2? 

To test this possibility, we create a simple recursive corpus: a \textit{Nesting Parentheses corpus} of hierarchically nesting matching symbols, and run the same experimental setup as we did for Experiments 1 and 2 \footnote{Though these corpora do not strictly use parentheses tokens, we refer to both of these as parentheses corpora, drawing our metaphor from the wide variety of studies such as \citet{karpathy16} examining nested parentheses.}. We find that plain recursion, even when the corpus has no other structural subtleties, is indeed a sufficient condition for inducing the kinds of structural transfer we observed in Experiment 2. 

Recursion is a sufficient quality, but is it the only explanation for our results? We also create a control corpus: a \textit{Flat Parentheses corpus}, which has similar pairs of matching parentheses, but which do not nest hierarchically and projectively (the difference between the two corpora is visually illustrated in Figure \ref{nonlingexamples}). We surprisingly find that this non-recursive corpus induces the same amount of structural transfer as the recursive nesting parentheses, which emphasizes the importance of pairing, head-dependency type structure in the linguistic structural embeddings of LSTMs.


\subsection{Data}

 The vocabulary for these corpora are the integers 0-50,000, where each number is a parenthesis token, and that token ``closes'' when the same integer appears a second time. We draw the opening tokens from the empirical Spanish unigram distribution (mapping each Spanish word to an integer), meaning that these corpora have a similar vocabulary distribution, albeit a much simpler non-linguistic structure, to the L2. Both of the corpora are 100 million tokens long, like the random and the natural language corpora.

We create the Nesting Parentheses corpus by following a simple stack-based grammar. At timestep $t$, we flip a coin to decide whether to open a new parenthesis (with probability 0.4) or close the top parenthesis on the stack (with probability 0.6).\footnote{$P(open)$ has to be strictly less than 0.5, or else the tree depth is expected to grow infinitely.} If we are opening a new parenthesis, we sample an integer $x_{open}$ from the Spanish unigram distribution, write the integer $x_{open}$ at the corpus position $t$, and push $x_{open}$ onto the stack of open parentheses. If we are closing a parenthesis, we pop the top integer from the stack, $x_{close}$, and write $x_{close}$ at corpus position $t$.

The Flat Parentheses corpus is made up of pairs of parentheses that do not nest. At timestep $t$, we sample an integer $x$ from the empirical Spanish unigram distribution, and a distance $d$ from the empirical distribution of dependency lengths (calculated from the Spanish Universal Dependencies treebank \cite{ud}). Then, we write $x$ at position $t$ and at position $t + d$. This creates pairs of matching parentheses which are not influenced by any other token in determining when they close. \textit{Note that this corpus is very similar to the Random Zipf corpus, except that each sampled token is placed twice instead of once.}

\subsection{Results}

LSTMs trained on both parentheses corpora are able to model human language far better than models trained on the random corpora, indicating that the isolated forms of grammar-like structure in these corpora are useful for modelling human language. Surprisingly, performance is the same for a model pretrained on the Nesting Parentheses and the Flat Parentheses corpus. This suggests that it is not necessarily hierarchical encodings which LSTMs use to model human language, and that other forms of structure such as flat head-head dependencies may be just as important \cite{marneffe19}.

The Nesting Parentheses corpus exhibits hierarchical structure while not having any of the irregularities and subtleties of human language or music. Despite the simplicity of the grammar, our results indicate that the presence of this hierarchical structure is very helpful for an LSTM attempting to model Spanish. Our models trained on the Nesting Parentheses corpus have an average perplexity of 170.98 when tested on the Spanish corpus. This is 322 perplexity points better than the baseline models trained on the Zipf Random corpus, which has the same vocabulary distribution (Figure \ref{nonlingresults}).

Models trained on the Flat Parentheses corpus are equally effective when tested on Spanish, achieving an average perplexity of 170.03. These results are surprising, especially given that the Flat Parentheses corpus is so similar to the Random Zipf corpus -- the only difference being that integers are placed in pairs not one by one -- and yet performs better by an average of 323 perplexity points. This suggests that \textit{representing relationships between pairs of tokens is a key element that makes syntactic representations of language successful in LSTMs}.

The Flat Parentheses corpus has structure in that each token is placed in relation to one other token, but just one other token. To model this successfully a model would have to have some ability to look back at previous tokens and determine which ones would likely have their match appear next. Our results suggest that this kind of ability is just as useful as potentially being able to model a simple stack-based grammar.

\section{Experiment 4: Human Languages}

    \begin{table}[t]
    \small
    \centering{
    \begin{tabular}{|l|c|}
    \hline
    Language&\parbox[t]{3cm}{WALS-syntax distance from Spanish (out of a max of 49 features)} \\
    \hline
    Spanish (es)&0 \\
    Italian (it)&0 \\
    Portuguese (pt)&3 \\
    English (en)&4 \\
    Romanian (ro)&5 \\
    Russian (ru)&9 \\
    German (de)&10 \\
    Finnish (fi)&13 \\
    Basque (eu)&15 \\
    Korean (ko)&18 \\
    Turkish (tr)&23 \\
    Japanese (ja)&23 \\
    \hline
    \end{tabular}
    }
    \caption{WALS-syntax distance between Spanish and L1s}
    \label{walstable}
    \end{table}

To further analyze what kinds of generalizable structure LSTMs can infer, we run experiments in transferring zero-shot between human languages. We ask: can LSTMs infer and use fine-grained syntactic similarities between typologically similar languages? Previous work \cite{zoph2016transfer, artetxezero} indicates that transfer is more successful between related languages. We control for vocabulary overlap, and use typological syntactic difference as a quantitative probe to ask: are fine-grained syntactic similarities encoded in generalizable, transferrable ways? To answer this question, we investigate the extent to which fine-grained differences in syntactic structure cause different zero-shot transfer results.

    \begin{figure}[t]
    \includegraphics[width=1.3\columnwidth]{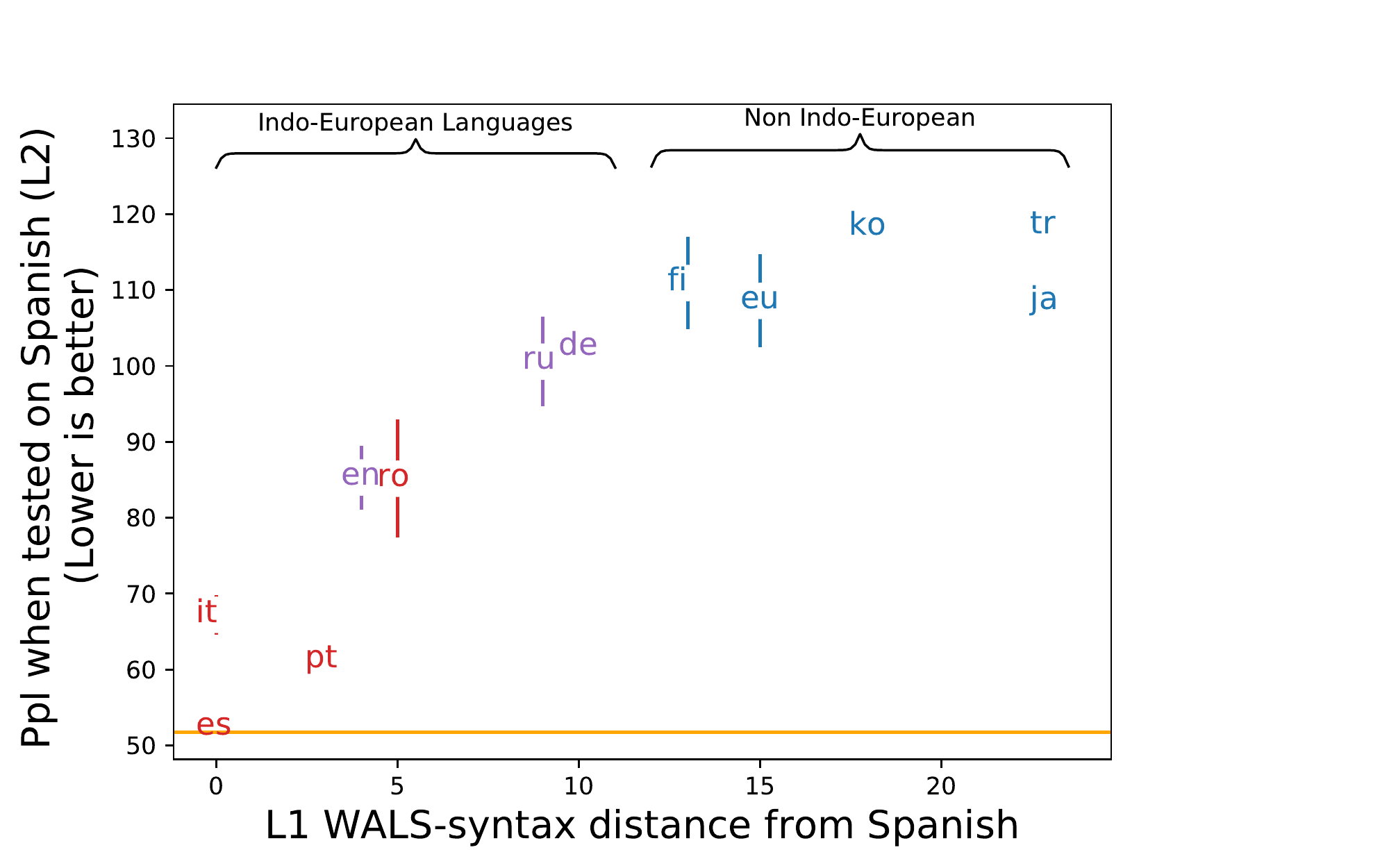}
    \caption{Results of Experiment 4. Transfer is better between typologically similar languages, even when vocabularies are disjoint. Perplexity on Spanish test data plotted against the WALS-syntax distance of each model's L1 to Spanish. The relationship is almost linear for Indo-European languages, and then reaches a ceiling. Error bars show 95\% CIs for $n=5$ trials with different random seeds. These results demonstrate how LSTMs can transfer knowledge more easily to languages that share structural features with the L1, and that this correlation is robust to multiple trials. The orange line represents the oracle perplexity of training all parameters to convergence on the L2 train data. Romance languages are in red, other Indo-European languages are in purple, and non-Indo-European languages are blue.}
    \label{walsscat}
    \end{figure}

\subsection{Data}

We created our language corpora from Wikipedia, which offers both wide language variation as well as a generally consistent tone and subject domain. We used the gensim wikicorpus library to strip Wikipedia formatting, and the stanfordnlp Python library \cite{stanfordnlp} to tokenize the corpus. We run experiments on data from 12 human languages, all of which have Wikipedias of over 100,000 articles: Spanish, Portuguese, Italian, Romanian, English, Russian, German, Finnish, Basque, Korean, Turkish and Japanese. All of the training corpora are 100 million tokens in length. \footnote{The code for recreating our corpora from Wikipedia dumps is available at \url{https://github.com/toizzy/wiki-corpus-creator}}

For our typological data, we use the World Atlas of Linguistic Structure, using the features that relate to syntax (WALS-syntax features). Examples of syntactic features in WALS include questions such as does a language have Subject-Verb-Object order, or does a degree word (like ``very'') come before or after the adjective.
We accessed the WALS data using the lang2vec package \cite{lang2vec}. The quantity we are interested in extracting from the WALS data is the \textit{typological distance} between the L2 (Spanish) and all of the L1 languages mentioned above. Not every feature is reported for every language, so we calculate the WALS distance by taking into account only the 49 (syntactic) features that are reported for all our chosen languages and count the number of entries that are different (see Table \ref{walstable}). Since they are only based on 49 features, these distances do not provide a perfectly accurate distance metric. Though we cannot use it for fine-grained analysis, correlation with this distance metric would imply correlation with syntactic distance.

\subsection{Results}

Our experiments present  a strong correlation between the ability to transfer from an L1 language to Spanish and the WALS-syntax distance between those two languages, as shown in Figure \ref{walsscat}(a). In the case of Indo-European languages the relationship is largely linear with a Pearson $R^2$ coefficient of 0.83. For languages not in the Indo-European language family, transfer performance appears to reach a noisy ceiling, and Pearson's $R^2=0.78$ when taking into account all languages.\footnote{We verified that our results also stand when calculating correlation coefficients using log perplexity, which yielded similar values: $R^2$ of 0.79 and 0.73 for Indo-European and all languages respectively.}

Our previous experiments show that LSTMs can encode and generalize structural features from data that is structured, both in recursive and in non-hierarchical fashion.  This experiment provides a more fine-grained analysis using using natural language to show that the syntax induced by LSTMs is generalizable to other languages in a typologically sensible fashion, even when we do not let the model take advantage of vocabulary overlap. However, after a certain threshold, the model is unable to take advantage of fine-grained similarities and performance on distant languages reaches a ceiling. It should be noted that all of the models trained on natural language, even the most distant, perform far better than non-linguistic data, indicating that LSTMs are able to extract universal syntactic information from all natural language L1s that is applicable to Spanish.


\section{Discussion}

In this work we propose the Test for Inductive bias via Language model Transfer (TILT), a novel analytic method for neural language models which tests the ability of a model to \textit{generalize and use structural knowledge}. We pretrain LSTMs on structured data, and then use the frozen LSTM weights to model human language. In doing so, we treat the frozen LSTM weights as the only structural faculty available to a human language model, and assess if the induced structure is general enough to be used to model human language.

Our experiments are cross-lingual and cross-modal in nature, not searching for representations of high-level features in one language, but for representations that encode general ideas of structure. While the majority of past work analyzing the structural abilities of neural models looks at a model's treatment of structural features that are realized in specific input sentences, our method compares the \textit{encoding and transfer of general grammatical features of different languages}. By using TILTs, we do not have to identify a structural feature of interest and investigate if it is being encoded, but instead asses if generalizable abstract structures are encoded in one language by examining \textit{if they can be used} to model human language. 
Our work thus avoids known issues that have been pointed out with analytic methods like  probing \cite{voitamdl, pimentel, johnpercy}.

We run experiments on  natural languages, artificial languages,  and non-linguistic corpora.
Our non-linguistic and artificial language experiments suggest three facets of the structural encoding ability of LSTM LMs. First, that vocabulary distribution has a very minor effect for modelling human language compared to structural similarity. Second, that models can encode useful language modelling information from the latent structure inherent in non-linguistic structured data, even if the surface forms are vastly differing. Last, that encodings derived from hierarchically structured tokens are equally useful for modelling human language as those derived from texts made up of pairs of tokens that are linked but non-hierarchical.
Running experiments on a range of human languages, we conclude that the internal linguistic representation of LSTM LMs allows them to take advantage of structural similarities between languages even when unaided by lexical overlap.

Our results on the parentheses corpora do not necessarily provide proof that the LSTMs trained on the Nesting Parentheses corpus aren't encoding and utilizing hierarchical structure. In fact, previous research shows that LSTMs are able to successfully model stack-based hierarchical languages \cite{suzgundyck, yudyck, suzguncounting}. What our results do indicate is that, in order for LSTMs to model human language, being able to model hierarchical structure is similar in utility to having access to a non-hierarchical ability to ``look back'' at one relevant dependency. These results shine light on the importance of considering other types of structural awareness that may be used by neural natural language models, even if those same models also demonstrate the ability to model pure hierarchical structure. 

Our method could be used to test many other hypotheses regarding neural language models, by choosing a discerning set of pretraining languages. A first step in future work would be to test if the results of this paper hold on Transformer architectures, or if instead  Transformers result in different patterns of structural encoding transfer.
Future work expanding on our results could focus on ablating specific structural features by creating hypothetical languages that differ in single grammatical features from the L2, in the style of Galactic Dependencies \cite{galactic}, and testing the effect of structured data that's completely unrelated to language, such as images.

Our results also contribute to the long-running nature-nurture debate in language acquisition:  whether the success of neural models implies that  unbiased learners can learn natural languages with enough data, or whether human abilities to acquire language given sparse stimulus implies a strong innate human learning bias \cite{linzenbaroni20}. The results of our parentheses experiments suggest that simple structural head-dependent bias, which need not be hierarchical, goes a long way toward making language acquisition possible for neural networks, highlighting the possibility of a less central role for recursion in language learning for both humans and machines.

\section*{Acknowledgements}

We thank Urvashi Khandelwal, Kawin Ethayarajh, Kyle Mahowald, Chris Donahue, Yiwei Luo, Alex Tamkin and Kevin Clark for helpful discussions and comments on drafts, and our anonymous reviewers for their feedback. 
This work was supported by an NSF Graduate Research Fellowship for IP and a SAIL-Toyota Research Award. Toyota Research Institute (``TRI'')  provided funds to assist the authors with their research but this article solely reflects the opinions and conclusions of its authors and not TRI or any other Toyota entity.
\bibliography{emnlp2020}

\begin{thebibliography}{32}
\expandafter\ifx\csname natexlab\endcsname\relax\def\natexlab#1{#1}\fi

\bibitem[{Artetxe et~al.(2020)Artetxe, Ruder, and Yogatama}]{artetxezero}
Mikel Artetxe, Sebastian Ruder, and Dani Yogatama. 2020.
\newblock \href {https://doi.org/10.18653/v1/2020.acl-main.421} {On the
  cross-lingual transferability of monolingual representations}.
\newblock In \emph{Proceedings of the 58th Annual Meeting of the Association
  for Computational Linguistics}, pages 4623--4637, Online. Association for
  Computational Linguistics.

\bibitem[{Clark et~al.(2019)Clark, Khandelwal, Levy, and Manning}]{urvashibert}
Kevin Clark, Urvashi Khandelwal, Omer Levy, and Christopher~D Manning. 2019.
\newblock What does {BERT} look at? an analysis of {BERT}’s attention.
\newblock In \emph{Proceedings of the 2019 ACL Workshop BlackboxNLP: Analyzing
  and Interpreting Neural Networks for NLP}, pages 276--286.

\bibitem[{Conneau et~al.(2018{\natexlab{a}})Conneau, Kruszewski, Lample,
  Barrault, and Baroni}]{conneaufing}
Alexis Conneau, German Kruszewski, Guillaume Lample, Loïc Barrault, and Marco
  Baroni. 2018{\natexlab{a}}.
\newblock \href {https://doi.org/10.18653/v1/p18-1198} {What you can cram into
  a single vector: Probing sentence embeddings for linguistic properties}.
\newblock \emph{Proceedings of the 56th Annual Meeting of the Association for
  Computational Linguistics (Volume 1: Long Papers)}.

\bibitem[{Conneau et~al.(2018{\natexlab{b}})Conneau, Rinott, Lample, Williams,
  Bowman, Schwenk, and Stoyanov}]{xnli}
Alexis Conneau, Ruty Rinott, Guillaume Lample, Adina Williams, Samuel Bowman,
  Holger Schwenk, and Veselin Stoyanov. 2018{\natexlab{b}}.
\newblock \href {https://doi.org/10.18653/v1/D18-1269} {{XNLI}: Evaluating
  cross-lingual sentence representations}.
\newblock In \emph{Proceedings of the 2018 Conference on Empirical Methods in
  Natural Language Processing}, pages 2475--2485, Brussels, Belgium.
  Association for Computational Linguistics.

\bibitem[{Dalvi et~al.(2019)Dalvi, Durrani, Sajjad, Belinkov, Bau, and
  Glass}]{grain}
Fahim Dalvi, Nadir Durrani, Hassan Sajjad, Yonatan Belinkov, Anthony Bau, and
  James Glass. 2019.
\newblock What is one grain of sand in the desert? analyzing individual neurons
  in deep {NLP} models.
\newblock In \emph{Proceedings of the AAAI Conference on Artificial
  Intelligence}, volume~33, pages 6309--6317.

\bibitem[{Gulordava et~al.(2018)Gulordava, Bojanowski, Grave, Linzen, and
  Baroni}]{gulordava}
Kristina Gulordava, Piotr Bojanowski, Edouard Grave, Tal Linzen, and Marco
  Baroni. 2018.
\newblock \href {https://doi.org/10.18653/v1/n18-1108} {Colorless green
  recurrent networks dream hierarchically}.
\newblock \emph{Proceedings of the 2018 Conference of the North American
  Chapter of the Association for Computational Linguistics: Human Language
  Technologies, Volume 1 (Long Papers)}.

\bibitem[{Hauser et~al.(2002)Hauser, Chomsky, and Fitch}]{hauser2002faculty}
Marc~D. Hauser, Noam Chomsky, and W.~Tecumseh Fitch. 2002.
\newblock \href {https://doi.org/10.1126/science.298.5598.1569} {The faculty of
  language: What is it, who has it, and how did it evolve?}
\newblock \emph{Science}, 298(5598):1569--1579.

\bibitem[{Hawthorne et~al.(2018)Hawthorne, Stasyuk, Roberts, Simon, Huang,
  Dieleman, Elsen, Engel, and Eck}]{maestro}
Curtis Hawthorne, Andriy Stasyuk, Adam Roberts, Ian Simon, Cheng-Zhi~Anna
  Huang, Sander Dieleman, Erich Elsen, Jesse Engel, and Douglas Eck. 2018.
\newblock \href {http://arxiv.org/abs/1810.12247} {Enabling factorized piano
  music modeling and generation with the maestro dataset}.

\bibitem[{Hewitt and Liang(2019)}]{johnpercy}
John Hewitt and Percy Liang. 2019.
\newblock Designing and interpreting probes with control tasks.
\newblock In \emph{Proceedings of the 2019 Conference on Empirical Methods in
  Natural Language Processing and the 9th International Joint Conference on
  Natural Language Processing (EMNLP-IJCNLP)}, pages 2733--2743.

\bibitem[{Hewitt and Manning(2019)}]{johnprobe}
John Hewitt and Christopher~D Manning. 2019.
\newblock A structural probe for finding syntax in word representations.
\newblock In \emph{Proceedings of the 2019 Conference of the North American
  Chapter of the Association for Computational Linguistics: Human Language
  Technologies, Volume 1 (Long and Short Papers)}, pages 4129--4138.

\bibitem[{Karpathy et~al.(2016)Karpathy, Johnson, and Fei-Fei}]{karpathy16}
Andrej Karpathy, Justin Johnson, and Li~Fei-Fei. 2016.
\newblock Visualizing and understanding recurrent networks.

\bibitem[{Lerdahl and Jackendoff(1996)}]{generativemusic}
Fred Lerdahl and Ray~S Jackendoff. 1996.
\newblock \emph{A generative theory of tonal music}.
\newblock MIT press.

\bibitem[{Linzen and Baroni(2020)}]{linzenbaroni20}
Tal Linzen and Marco Baroni. 2020.
\newblock Syntactic structure from deep learning.
\newblock \emph{Annual Review of Linguistics}, 7.

\bibitem[{Linzen et~al.(2016)Linzen, Dupoux, and Goldberg}]{linzen16}
Tal Linzen, Emmanuel Dupoux, and Yoav Goldberg. 2016.
\newblock Assessing the ability of {LSTM}s to learn syntax-sensitive
  dependencies.
\newblock \emph{Transactions of the Association for Computational Linguistics},
  4:521--535.

\bibitem[{Littell et~al.(2017)Littell, Mortensen, Lin, Kairis, Turner, and
  Levin}]{lang2vec}
Patrick Littell, David~R Mortensen, Ke~Lin, Katherine Kairis, Carlisle Turner,
  and Lori Levin. 2017.
\newblock Uriel and lang2vec: Representing languages as typological,
  geographical, and phylogenetic vectors.
\newblock In \emph{Proceedings of the 15th Conference of the European Chapter
  of the Association for Computational Linguistics: Volume 2, Short Papers},
  volume~2, pages 8--14.

\bibitem[{de~Marneffe and Nivre(2019)}]{marneffe19}
Marie-Catherine de~Marneffe and Joakim Nivre. 2019.
\newblock Dependency grammar.
\newblock \emph{Annual Review of Linguistics}, 5:197--218.

\bibitem[{McCoy et~al.(2020)McCoy, Frank, and Linzen}]{mccoy2020}
R.~Thomas McCoy, Robert Frank, and Tal Linzen. 2020.
\newblock \href {https://doi.org/10.1162/tacl_a_00304} {Does syntax need to
  grow on trees? sources of hierarchical inductive bias in sequence-to-sequence
  networks}.
\newblock \emph{Transactions of the Association for Computational Linguistics},
  8:125–140.

\bibitem[{McDonald et~al.(2013)McDonald, Nivre, Quirmbach-Brundage, Goldberg,
  Das, Ganchev, Hall, Petrov, Zhang, T{\"a}ckstr{\"o}m et~al.}]{ud}
Ryan McDonald, Joakim Nivre, Yvonne Quirmbach-Brundage, Yoav Goldberg, Dipanjan
  Das, Kuzman Ganchev, Keith Hall, Slav Petrov, Hao Zhang, Oscar
  T{\"a}ckstr{\"o}m, et~al. 2013.
\newblock Universal dependency annotation for multilingual parsing.
\newblock In \emph{Proceedings of the 51st Annual Meeting of the Association
  for Computational Linguistics (Volume 2: Short Papers)}, pages 92--97.

\bibitem[{Merity et~al.(2018)Merity, Keskar, and Socher}]{awdlm}
Stephen Merity, Nitish~Shirish Keskar, and Richard Socher. 2018.
\newblock \href {https://openreview.net/forum?id=SyyGPP0TZ} {Regularizing and
  optimizing {LSTM} language models}.
\newblock In \emph{International Conference on Learning Representations}.

\bibitem[{Movshovitz-Attias and Cohen(2013)}]{habeas}
Dana Movshovitz-Attias and William Cohen. 2013.
\newblock Natural language models for predicting programming comments.
\newblock In \emph{Proceedings of the 51st Annual Meeting of the Association
  for Computational Linguistics (Volume 2: Short Papers)}, pages 35--40.

\bibitem[{Piantadosi(2014)}]{piantadosizipf}
Steven~T Piantadosi. 2014.
\newblock Zipf’s word frequency law in natural language: A critical review
  and future directions.
\newblock \emph{Psychonomic bulletin \& review}, 21(5):1112--1130.

\bibitem[{Pimentel et~al.(2020)Pimentel, Valvoda, Maudslay, Zmigrod, Williams,
  and Cotterell}]{pimentel}
Tiago Pimentel, Josef Valvoda, Rowan~Hall Maudslay, Ran Zmigrod, Adina
  Williams, and Ryan Cotterell. 2020.
\newblock \href {http://arxiv.org/abs/2004.03061} {Information-theoretic
  probing for linguistic structure}.

\bibitem[{Ponti et~al.(2019)Ponti, Vuli{\'c}, Cotterell, Reichart, and
  Korhonen}]{edo-zero}
Edoardo~Maria Ponti, Ivan Vuli{\'c}, Ryan Cotterell, Roi Reichart, and Anna
  Korhonen. 2019.
\newblock \href {https://doi.org/10.18653/v1/D19-1288} {Towards zero-shot
  language modeling}.
\newblock In \emph{Proceedings of the 2019 Conference on Empirical Methods in
  Natural Language Processing and the 9th International Joint Conference on
  Natural Language Processing (EMNLP-IJCNLP)}, pages 2900--2910, Hong Kong,
  China. Association for Computational Linguistics.

\bibitem[{Qi et~al.(2018)Qi, Dozat, Zhang, and Manning}]{stanfordnlp}
Peng Qi, Timothy Dozat, Yuhao Zhang, and Christopher~D. Manning. 2018.
\newblock \href {https://nlp.stanford.edu/pubs/qi2018universal.pdf} {Universal
  dependency parsing from scratch}.
\newblock In \emph{Proceedings of the {CoNLL} 2018 Shared Task: Multilingual
  Parsing from Raw Text to Universal Dependencies}, pages 160--170, Brussels,
  Belgium. Association for Computational Linguistics.

\bibitem[{Suzgun et~al.(2019{\natexlab{a}})Suzgun, Gehrmann, Belinkov, and
  Shieber}]{suzguncounting}
Mirac Suzgun, Sebastian Gehrmann, Yonatan Belinkov, and Stuart~M. Shieber.
  2019{\natexlab{a}}.
\newblock \href {https://www.aclweb.org/anthology/W19-3900} {{LSTM} networks
  can perform dynamic counting}.
\newblock In \emph{Proceedings of the Workshop on Deep Learning and Formal
  Languages: Building Bridges}, volume abs/1906.03648, Florence. Association
  for Computational Linguistics.

\bibitem[{Suzgun et~al.(2019{\natexlab{b}})Suzgun, Gehrmann, Belinkov, and
  Shieber}]{suzgundyck}
Mirac Suzgun, Sebastian Gehrmann, Yonatan Belinkov, and Stuart~M. Shieber.
  2019{\natexlab{b}}.
\newblock \href {http://arxiv.org/abs/1911.03329} {Memory-augmented recurrent
  neural networks can learn generalized {D}yck languages}.

\bibitem[{Talmor et~al.(2019)Talmor, Elazar, Goldberg, and Berant}]{olmpics}
Alon Talmor, Yanai Elazar, Yoav Goldberg, and Jonathan Berant. 2019.
\newblock \href {http://arxiv.org/abs/1912.13283} {o{LM}pics -- on what
  language model pre-training captures}.

\bibitem[{Voita and Titov(2020)}]{voitamdl}
Elena Voita and Ivan Titov. 2020.
\newblock Information-theoretic probing with minimum description length.
\newblock \emph{arXiv preprint arXiv:2003.12298}.

\bibitem[{Wang and Eisner(2016)}]{galactic}
Dingquan Wang and Jason Eisner. 2016.
\newblock The {G}alactic {D}ependencies treebanks: Getting more data by
  synthesizing new languages.
\newblock \emph{Transactions of the Association for Computational Linguistics},
  4:491--505.

\bibitem[{Wu and Dredze(2019)}]{beto}
Shijie Wu and Mark Dredze. 2019.
\newblock \href {https://doi.org/10.18653/v1/d19-1077} {Beto, bentz, becas: The
  surprising cross-lingual effectiveness of bert}.
\newblock \emph{Proceedings of the 2019 Conference on Empirical Methods in
  Natural Language Processing and the 9th International Joint Conference on
  Natural Language Processing (EMNLP-IJCNLP)}.

\bibitem[{Yu et~al.(2019)Yu, Vu, and Kuhn}]{yudyck}
Xiang Yu, Ngoc~Thang Vu, and Jonas Kuhn. 2019.
\newblock Learning the dyck language with attention-based seq2seq models.
\newblock In \emph{ACL 2019}.

\bibitem[{Zoph et~al.(2016)Zoph, Yuret, May, and Knight}]{zoph2016transfer}
Barret Zoph, Deniz Yuret, Jonathan May, and Kevin Knight. 2016.
\newblock \href {https://doi.org/10.18653/v1/D16-1163} {Transfer learning for
  low-resource neural machine translation}.
\newblock In \emph{Proceedings of the 2016 Conference on Empirical Methods in
  Natural Language Processing}, pages 1568--1575, Austin, Texas. Association
  for Computational Linguistics.

\end{thebibliography}
\bibliographystyle{acl_natbib}

\section*{Appendix: Numerical results of experiments}
For every experiment we ran five trials with different random seeds. We list the means and standard deviations for each L1 below:

\begin{table}[h]
\begin{tabular}{|rlr|}
  \hline
  L1 Language & Mean TILT Ppl & Std. Dev \\ 
  \hline
  Random Uniform & 513.66 & 1.01 \\ 
  Random Zipf & 493.15 & 2.97 \\ 
  \hline 
  Music & 256.15 & 2.65 \\ 
  Code & 139.11 & 1.24 \\ 
  \hline
  Nesting Parens & 170.98 & 1.02 \\ 
  Flat Parens & 170.30 & 1.48 \\
  \hline
  Basque & 108.57 & 4.93 \\ 
  English & 85.30 & 3.40 \\ 
  Finnish & 110.92 & 3.84 \\ 
  German & 102.42 & 0.51 \\ 
  Italian & 67.21 & 2.10 \\ 
  Japanese & 108.48 & 0.81 \\ 
  Korean & 118.23 & 1.23 \\ 
  Portoguese & 61.25 & 0.21 \\ 
  Romanian & 85.14 & 6.26 \\ 
  Russian & 100.56 & 4.74 \\ 
  Spanish & 52.33 & 0.21 \\ 
  Turkish & 118.45 & 0.85 \\ 
   \hline
\end{tabular}
\end{table}

\end{document}